\documentclass[10pt,twocolumn,letterpaper]{article}

\usepackage{cvpr}              

\usepackage{graphicx}
\usepackage{amsmath}
\usepackage{amssymb}
\usepackage{booktabs}
\usepackage{multirow}
\usepackage{multicol}
\usepackage{color, colortbl}
\usepackage{pifont}
\newcommand{\cmark}{\ding{51}}%
\usepackage[pagebackref,breaklinks,colorlinks]{hyperref}

\usepackage[capitalize]{cleveref}
\crefname{section}{Sec.}{Secs.}
\Crefname{section}{Section}{Sections}
\Crefname{table}{Table}{Tables}
\crefname{table}{Tab.}{Tabs.}

\def\cvprPaperID{*****} 
\def\confName{CVPR}
\def\confYear{2023}

\begin{document}
\pagenumbering{arabic}
\renewcommand\thefootnote{}
\title{Imitation with Spatial-Temporal Heatmap:  \\ 2nd Place Solution for NuPlan Challenge}
\author{
Yihan Hu\thanks{Equal contribution.}, Kun Li\footnotemark[1], Pingyuan Liang\footnotemark[1],
Jingyu Qian, Zhening Yang, \\ 
Haichao Zhang, Wenxin Shao, Zhuangzhuang Ding\thanks{Work done while at Horizon Robotics.}, Wei Xu, Qiang Liu \\
Horizon Robotics \\
{\tt\small \{yihan.hu96, kun.li0929\}@gmail.com}
}

\maketitle

\begin{abstract}
    This paper presents our 2nd place solution for the NuPlan Challenge 2023. Autonomous driving in real-world scenarios is highly complex and uncertain. Achieving safe planning in the complex multimodal scenarios is a highly challenging task. Our approach, Imitation with Spatial-Temporal Heatmap, adopts the learning form of behavior cloning, innovatively predicts the future multimodal states with a heatmap representation, and uses trajectory refinement techniques to ensure final safety. The experiment shows that our method effectively balances the vehicle's progress and safety, generating safe and comfortable trajectories. In the NuPlan competition, we achieved the second highest overall score, while obtained the best scores in the ego progress and comfort metrics.
\end{abstract}

\section{Introduction}
\label{sec:intro}
The NuPlan challenge is the world's first large-scale planning benchmark for autonomous driving~\cite{nuplan}. It offers approximately 1300 hours of human driving data sourced from four different cities across the US and Asia with highly complex and diverse scenarios. Unlike previous competitions, NuPlan focuses more on long-term planning rather than short-term motion prediction. It also provides a highly realistic simulator for conducting closed-loop evaluation that aligns more closely with real-world scenarios. Amidst the numerous competitors, our methodology stood out as the second place in the competition.

We use an imitation learning approach with several innovations, including using a spatial-temporal heatmap to represent the distributions of the ego future trajectory, and using a multitasking learning approach to predict and model the surrounding dynamic objects. With these dynamic and static future environment, we find the optimal trajectory with a post-solver. Our ablation studies demonstrate the effectiveness of these techniques.

\begin{figure*}[h]
\centering
\includegraphics[width=0.8\textwidth]{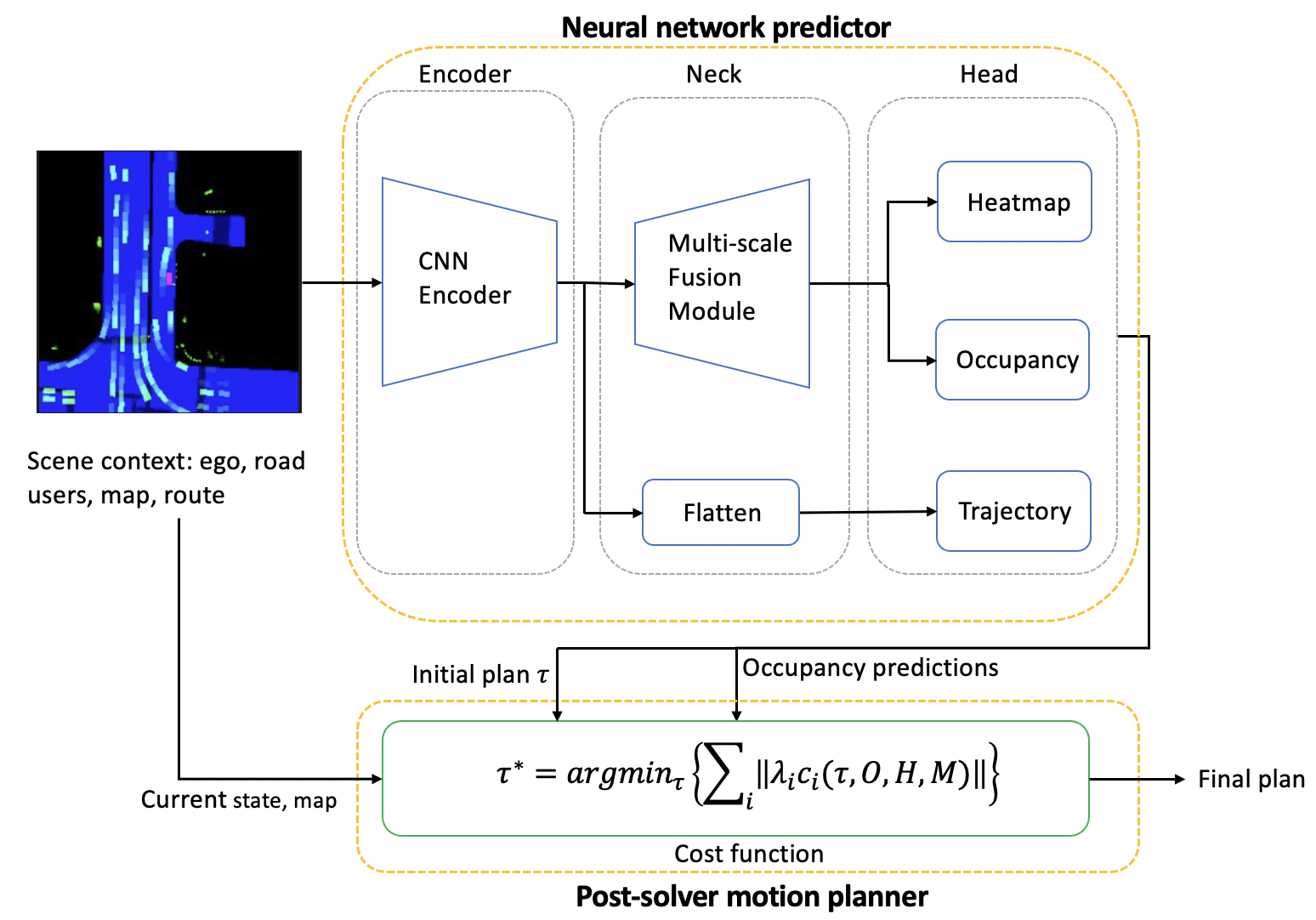} 
\caption{Overall structure of our model. The neural network predictor predicts the future occupancy of the surrounding agents and the initial plan. A post-solver is employed to explicitly refine the planned trajectory. $\tau$ denotes the plan trajectory of the ego vehicle, $\lambda_{\text{i}}$ denotes the weight hyperparameters associated with the cost function $c_{i}$, $O$ denotes as occupancy predictions, $H$ denotes the heatmap prediction, and $M$ denotes the input HD map.}
\label{fig:framework}
\end{figure*}

\section{Methodology}
In this section, we describe our model as shown in Fig.~\ref{fig:framework}. Our system comprises two stages: the behavior cloning stage and the trajectory refinement stage. In the first stage, the model is trained by supervised learning to generate the initial plan, multi-modal heatmap, and the future behavior of surrounding agents. The second stage focuses on refining the initial trajectory, taking into account the vehicle's kinematics, comfort, and safety constraints. We first introduce our input representation, and then elaborate the structure of our model, and lastly give a more detailed description of the trajectory refinement stage.

\subsection{Input Representation}


In the context of autonomous driving, raster serves as a grid-based representation of the environment. They offer a snapshot of the surroundings, where the grids contain essential information of road layout, traffic conditions, and vehicle status. Raster provides a structured format that is particularly useful for processing spatial domain information.


For this competition, we created a six-channel raster, each channel realizing an environmental modality, including ego vehicle's current state, other agents's history, current map information and navigation route, etc.. The ego channel represents our vehicle's state and position. The road map channel encapsulates physical layout features, transformed from a vector-based map. Baseline-path channel displays all lanes within a specific range. Agent channels encode traffic participants other than the ego car, using 2D representations on the raster grid. The route raster outlines a navigation route for the planning model to follow. Finally, the ego speed channel fills the 2D raster with the ego vehicle's speed. This multi-channel raster input offers a comprehensive spatio-temporal environmental depiction, enabling the model to comprehend and anticipate traffic dynamics.

\subsection{Network Structure}
\textbf{Encoder} To encode the input data, we employ ResNet~\cite{he2015deep} as our convolutional neural network (CNN) encoder. Specifically, we utilize ResNet50 in our implementation. The encoder is responsible for generating multi-scale features denoted as $C_1, C_2, C_3, C_4, C_5$, where $C_i$ represents the feature map with a spatial size of $\frac{H}{2^i} \times \frac{W}{2^i}$, serving as the input for the subsequent components.

\textbf{Neck} The neck component in our model follows the architecture of Unet~\cite{ronneberger2015unet}, which facilitates the integration of features at multiple resolutions, enabling the model to capture both fine-grained and high-level contextual information. 

\subsubsection{Heads}
Our method employs three distinct heads dedicated to different tasks, namely ego trajectory prediction, ego heatmap prediction and surrounding agents occupancy predictions.

\textbf{Trajectory head} The trajectory head generates the initial plan, which is composed of two fully-connected layers. The output trajectory is denoted as $\tau \in \mathbb{R}^{T \times 3}$, where $T$ represents the number of time steps and 3 denotes the three-dimensional parameters (e.g., x, y, heading) of the predicted trajectory.

\textbf{Heatmap head} 
Inspired by HOME~\cite{gilles2021home}, we adopt a bird-eye-view heatmap representation for ego trajectory. Differently, we model ego location at each time step. Each pixel in the output image represents a location on the ground, and the value at each pixel indicates the probability or confidence associated with the presence of each trajectory point at that particular location. In order to predict the ego plan on a more finely-grained scale, we up-sample the original feature to 0.25m × 0.25m/pixel.

To generate the target output, we use a Gaussian distribution centered around the ground truth position. This approach allows the model to capture the uncertainty and multi-modality in the each trajectory point prediction.

\textbf{Occupancy head}  To model the surrounding dynamic environment, we employ the occupancy head, which predicts the motion behavior of other agents in the form of probabilistic occupancy grids~\cite{hu2022hope}. Specifically, the occupancy head predicts occupancy logits for each timestep t and shares the same feature resolution as the input raster.

\subsection{Collision Avoidance Map}
To better serve downstream applications, it is necessary to transform the predicted ego vehicle pose, the predicted occupancy probability of surrounding agents, and the static information (HD maps) into a collision probability density map. The collision probability density map represents the probability of collision at a given location in the coordinate system of the predicted future trajectory. Innovatively, we leverage the group convolution operator~\cite{chollet2017xception} on the GPU to efficiently execute this step and achieve real-time performance.
Specifically, we merge other agents' predicted occupancy $\hat{O}_\text{agents}$, static objects $\tilde{O}_\text{static}$, and the drivable area $\tilde{O}_\text{drivable}$ into a non-drivable area map $\hat{O}_{\text{non-drivable}}$. With the predicted ego vehicle pose $\hat{\tau_{t}}$, we create a convolution kernel $W(\hat{\tau_{t}}, H_{ego}, W_{ego})$ that matches the shape and future pose of the ego vehicle. By performing a convolution of this kernel on the non-drivable area, we can obtain the desired collision probability density map:
\begin{equation}
\label{eq:non-drivable}
\hat{O}_{\text{non-drivable}}^t = \hat{O}_\text{agent}^t|\tilde{O}_\text{static}|\overline{\tilde{O}}_\text{drivable},
\end{equation}
\vspace{-25pt}

\begin{equation}
\label{eq:colmap}
\hat{O}^t = \text{Conv}(W(\hat{\tau_{t}}, H_{ego}, W_{ego}), \tilde{O}_\text{non-drivable}).
\end{equation}
\subsection{Post-Solver}
Similar to UniAD~\cite{Hu_2023_CVPR}, we employ CasADi~\cite{Andersson2019} ipopt solver, which takes into account the vehicle's kinematics, comfort, predicted heatmap probability, and collision probability density map. By adjusting the initial trajectory through the post solver, we aim to achieve a safe and comfortable trajectory for vehicle control. Specifically, we denote the output trajectory as the parameter $\tau \in \mathbb{R}^{T \times 3}$, the imitated initial trajectory as $\hat{\tau} \in \mathbb{R}^{T \times 3}$, the collision probability density map as $\hat{O} \in \mathbb{R}^{h \times w \times T}$, the heatmap prediction as $\hat{H}\in \mathbb{R}^{h \times w \times T}$. The cost function $f(\cdot)$ is calculated by:
\begin{equation}
\label{eq:col-cost}
\begin{aligned}
    f(\tau|\hat{\tau}, \hat{O}, \hat{H}) = & \lambda_{\text{imi}} \lVert\tau, \hat{\tau}\rVert_2 
    +  \sum_{\phi\in \Phi} \phi(\tau) 
     + \lambda_{\text{o}}  \sum_{t} \mathcal{D}_{\text{o}}(\tau_{t}, \hat{O}^{t}) \\ & - \lambda_{\text{h}}  \sum_{t} \mathcal{D}_{\text{h}}(\tau_{t}, \hat{H}^{t}), 
\end{aligned}
\end{equation}
\vspace{-10pt}
\begin{equation}
\label{eq:col-dist}
    \mathcal{D}_{\text{o}}(\tau_{t}, \hat{O}^{t}) = \sum_{(x, y) \in \mathcal{S}_{\text{o}}} 
    \frac{\hat{O}^{t}(x, y)}{\sigma_{\text{o}} \sqrt{2\pi}}
    \text{exp} (-\frac{\lVert \tau_{t} - (x, y) \rVert_2^2}{2\sigma_{\text{o}}^2}),
\end{equation}
\begin{equation}
\label{eq:hm-dist}
    \mathcal{D}_{\text{h}}(\tau_{t}, \hat{H}^{t}) = \sum_{(x, y) \in \mathcal{S}_{\text{h}}} 
    \frac{\hat{H}^{t}(x,y)}{\sigma_{\text{h}} \sqrt{2\pi}}
    \text{exp} (-\frac{\lVert \tau_{t} - (x, y) \rVert_2^2}{2\sigma_{\text{h}}^2}),
\end{equation}
where $\lambda_{\text{imi}}$, $\lambda_{\text{o}}$, $\lambda_{\text{h}}$ are the hyperparameters, and the kinematic function set $\Phi$ has five terms including jerk, curvature, curvature rate, acceleration and lateral acceleration. To speed up the inference, we sample the $S_o$ nearest occupied pixels and the top $S_h$ heatmap pixels at each time step. Moreover, to ensure the model can output trajectories that are consistent with actual physical conditions, we add some hard constraints, including dynamic constraints for the ego vehicle, state constraints, and control constraints.

\begin{table*}[h]
\begin{center}
    \definecolor{Gray}{gray}{0.9}
    \newcolumntype{g}{>{\columncolor{Gray}}c}
    \resizebox{\textwidth}{!}{
        \begin{tabular}{ccc|gccc|ccccccc}
        \hline
          perturbation & heatmap  & post-solver & final score & open loop & cl-nr & cl-r & collisions & TTC & drivable& comfort &	progress &	speed limit & direction 
						 \\
       \hline \hline
       & & & 0.782 & 0.825 & 0.773 & 0.748 & 0.878 & 0.817 & 0.985 & 0.991 & 0.985 & 0.969 & 0.985  \\
        \cmark & & & 0.803 & \textbf{0.865} & 0.786 & 0.758 & 0.886 & 0.835 & 0.951 & 0.988 & 0.922 & 0.959 & 0.987\\ 
        \cmark & \cmark & & 0.832 & 0.854 & 0.836 & 0.805 & 0.935 & 0.860 & 0.964 & 0.980 & \textbf{0.931} & 0.960 & 0.989 \\
        \cmark & \cmark & \cmark &  \textbf{0.876} & 0.852 & \textbf{0.896} & \textbf{0.882} & \textbf{0.967} & \textbf{0.910} & \textbf{0.994} & \textbf{0.992} & 0.920 & \textbf{0.962} & \textbf{0.992}\\
        \hline
        \end{tabular}
        }
\end{center}
\caption{Ablation Study. ``cl-nr" means close loop non-reactive simulation, which is identical to challenge 2 in the NuPlan challenge. ``cl-r" means close loop reactive simulation, which is identical to  challenge 3. In the last major column, we present the detailed metrics for close loop non-reactive simulation. ``TTC" means time-to-collision.} 
\label{tbl:ablation study}
\end{table*}

\begin{table*}[h]
\begin{center}
    \definecolor{Gray}{gray}{0.9}
    \newcolumntype{g}{>{\columncolor{Gray}}c}
    \resizebox{\textwidth}{!}{
        \begin{tabular}{cc|gccc|ccccccc}
        \hline
          rank & methods & final score & open loop (ch1) & cl-nr (ch2) & cl-r (ch3) & collisions & TTC & drivable& comfort &	progress &	speed limit & direction 
						 \\
       \hline  \hline
       1& 	CS\_Tu &  0.90 & 0.83 & 0.93 & 0.93 & 	0.99 & 	0.93 & 1.00 & 0.92 & 0.91 & 1.00 & 1.00 \\
        2 & \textbf{hoplan(Ours)} & 0.87 & 0.85 & 0.89 & 0.88 & 0.96 & 0.91 & 0.99 & \textbf{0.99} & \textbf{0.92} & 0.96 & 0.99\\ 
         3 & pegasus\_multipath & 0.85 & 0.88 & 0.82 & 0.85 & 	0.93 & 0.88 & 0.95 & 0.93 & 0.79 & 0.93 & 0.95 \\
         4 & GameFormer &  0.83 & 0.84 & 0.81 & 0.84 & 0.94 & 0.88 & 0.96 & 0.94 & 0.84 & 0.97 & 0.99\\
        \hline
        \end{tabular}
        }
\end{center}
\caption{Final leaderboard of the NuPlan Challenge on the private test set. Here we display the scores for challenge 1,2 and 3, and the final ranking was an average of these three scores, as indicated in the gray area of the table. In the last major column, we present the detailed metrics for close loop non-reactive simulation.} 
\label{tbl:leaderboard}
\end{table*}

\subsection{Learning}
We adopted a multi-task learning approach. For the prediction of the initial trajectory and pose of the ego vehicle, we employed a weight-decay L1 loss as shown in eq~\ref{eq:imitation}. For the heatmap supervision, we utilized the penalty-reduced pixelwise logistic regression with focal loss\cite{lin2017focal, hu2022afdetv2} as shown in eq~\ref{eq:hm}, where $\hat{H}_t$ is the predicted ego vehicles's location at time t, and $\tilde{H}_t$ is the target ground truth. Positive samples correspond to the locations of the expert trajectory, denoted as $\Tilde{\tau}$. All other locations are considered negative samples, with penalties attenuated by a Gaussian kernel. For the occupancy, we applied binary cross-entropy loss. And lastly we take a weighted sum as the final loss.
\begin{equation}
\label{eq:imitation}
    L_{\text{imi}} = \sum_t^T \text{exp}(\frac{t}{\alpha T}) \lVert \hat{\tau}_t, \tilde{\tau}_t\rVert_1
\end{equation}
\vspace{-10pt}
\begin{equation}
\label{eq:hm}
    L_{\text{hm}} = \sum_t \mathcal{H}(\hat{H}_t, \tilde{H}_t)
\end{equation}
\begin{equation}
\label{eq:occ}
    L_{\text{occ}} = \sum_t \mathcal{B}(\hat{O}_t, \tilde{O}_t)
\end{equation}
\begin{equation}
\label{eq:all-loss}
    L = \lambda_{\text{imi}} * L_{\text{imi}} + \lambda_{\text{hm}} * L_{\text{hm}} + \lambda_{\text{occ}} * L_{\text{occ}}.
\end{equation}

\section{Experiments}
We generated the rasterized input within a $112m \times 112m$ region at a resolution of 0.5m/pixel, resulting in an input spatial size of $224\times224$. To optimize our model, we employ the Adam optimizer~\cite{kingma2017adam} along with a multiple-step policy. The initial learning rate was set to $2\times 10^{-4}$, and a weight decay of $5\times10^{-4}$ was applied. Furthermore, we assign the following loss coefficients: $\lambda_{\text{occ}}=100$, $\lambda_{\text{hm}}=1.0$, and $\lambda_{\text{imi}}=1.0$. The model was trained for 20 epochs with a batch size of 32.

Regarding the data, we employed a random sampling strategy where 50,000 frames per scenario type were selected from the training dataset, resulting in a total of approximately 1.5 million frames.

To enhance the model's performance, we introduced perturbations during training, inspired by the methodology proposed by ChauffeurNet~\cite{bansal2018chauffeurnet}. Specifically, we applied a uniformly distributed random jittering to the current pose of the ego agent within the ranges of [0, 1.0] meters along the x-axis and [-1.0, 1.0] meters along the y-axis. Additionally, the heading was perturbed by an angle between [-0.25, 0.25] in radians. To ensure smooth trajectories, we fit a trajectory starting at the perturbed point and ended at the original end point, under a variety of dynamic constraints. These perturbed training examples enabled the ego car to recover its normal trajectory if experiencing a deviation from its normal route.

\section{Results}
\subsection{Ablation Study}
In this section, we conduct an ablation study on the aforementioned techniques, as shown in Tab.~\ref{tbl:ablation study}. We can observe that perturbation improved both open-loop and close-loop performances. As a form of data augmentation, perturbation can significantly enhance the data utilization rate. For the heatmap prediction, we saw a substantial enhancement in close-loop performance, particularly in close-loop reactive scenarios and collision rate. This demonstrates that the spatial-temporal heatmap could be a better representation of planning compared to single trajectory. In later visualizations in Sec.~\ref{sec:vis}, we can observe the effectiveness of the heatmap representation in modeling multi-modality and uncertainty. Furthermore, the bird's eye view spatial representation aligns well with our raster input, guiding the model's convergence. Furthermore, by incorporating the post-solver, we noticed a significant boost in close-loop performance, notably in collision and drivable area compliance, validating the effectiveness of post-optimization in the modeled environment.

\subsection{Learderboard}
Here, we present our ranking on the private test set in the NuPlan competition, as shown in Tab.~\ref{tbl:leaderboard}. Notably, We achieve the highest comfort and ego progress among all competitors.
\begin{figure*}[ht]
\begin{minipage}[b]{1\textwidth}
  \centering
  \includegraphics[width=\textwidth]{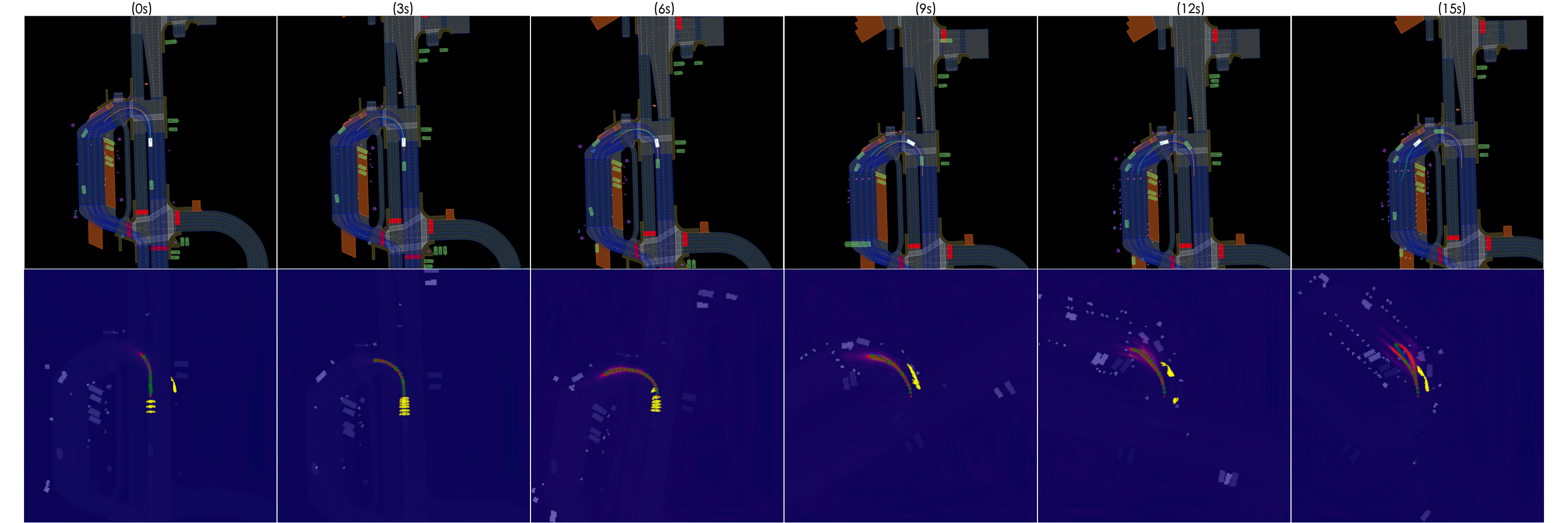}
  \caption{The ego vehicle expertly navigates a wide left turn, seamlessly entering the designated pick-up zone.}
  \label{fig:left_turn_vis}
\end{minipage}
\hfill
\begin{minipage}[b]{1\textwidth}
  \centering
  \includegraphics[width=\textwidth]{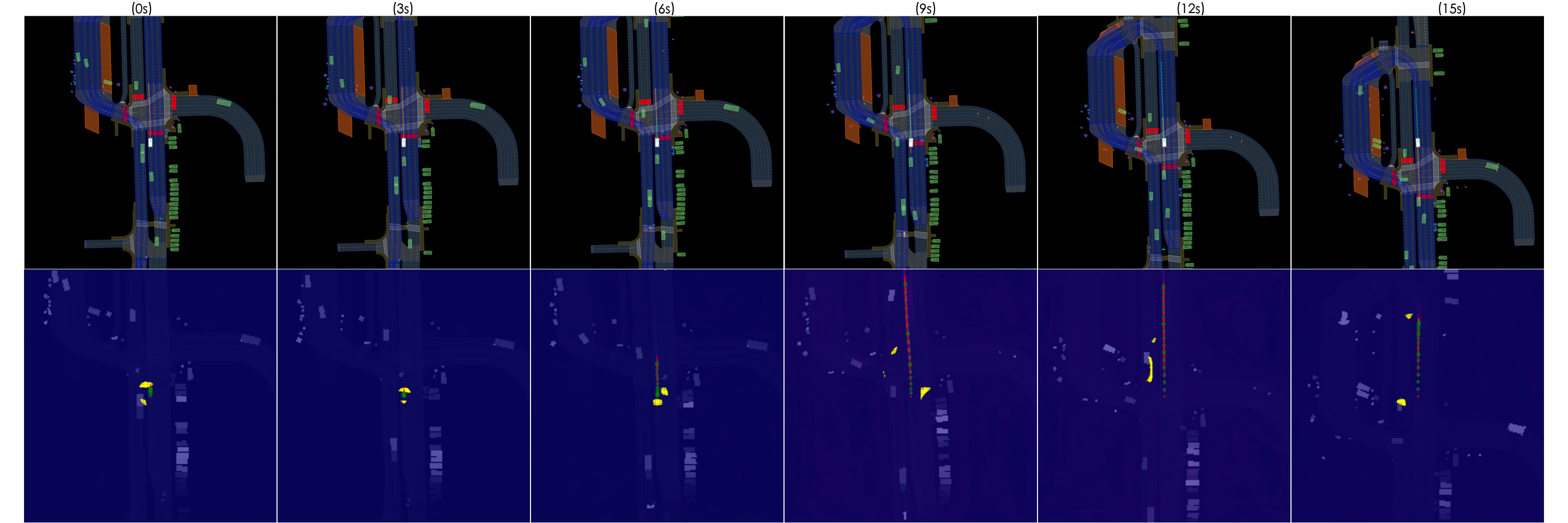}
  \caption{The ego vehicle exhibits patience as it allows several pedestrians to safely traverse the crosswalk, subsequently accelerating smoothly to continue its journey once the path is clear.}
  \label{fig:waiting_for_pedestrian_vis}
\end{minipage}
\hfill
\begin{minipage}[b]{1\textwidth}
  \centering
  \includegraphics[width=\textwidth]{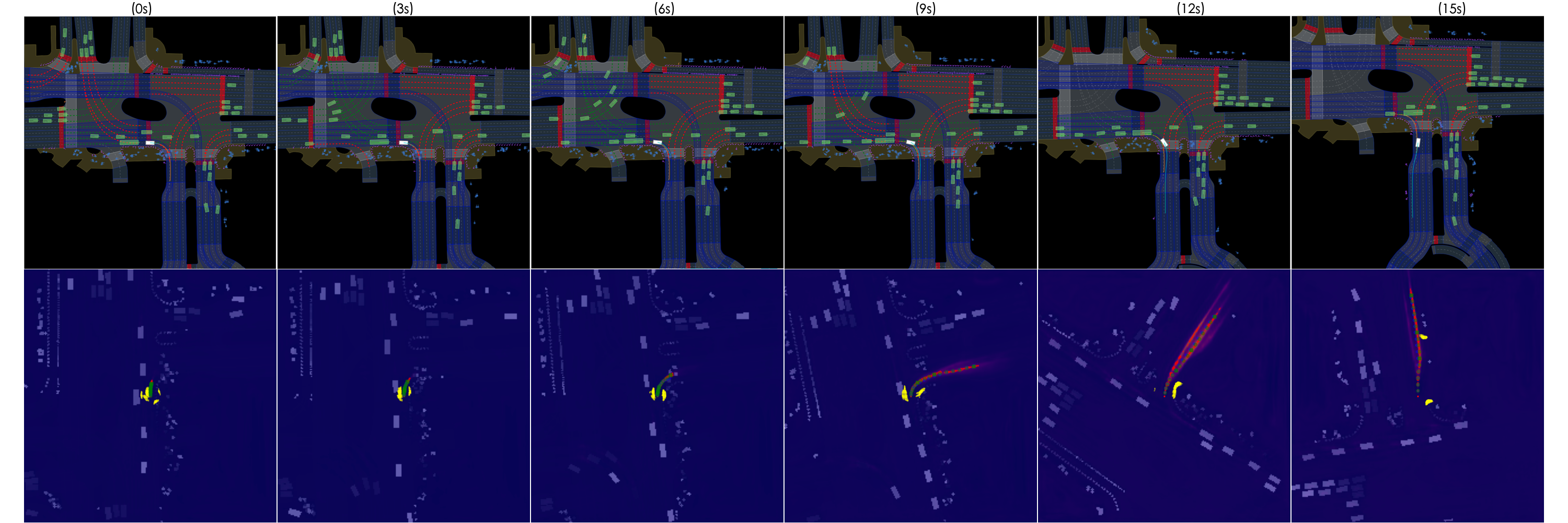}
  \caption{The ego vehicle demonstrates patience, awaiting numerous pedestrians to complete their crossing before executing a smooth, unprotected right turn with precision.}
  \label{fig:right_turn_vis}
\end{minipage}
\end{figure*}

\subsection{Visualizations}
\label{sec:vis}
In this section, we present a qualitative assessment of our planner's performance through closed-loop simulation results, under representative driving scenarios. These scenarios are visualized as sequential snapshots of the closed-loop rollouts. The top row displays images generated using the nuboard visualization tool, illustrating the smooth movement of the ego vehicle, denoted by a white rectangle. The bottom row demonstrates corresponding model predictions. Here, the heatmap predictions are indicated in red, pixels of potential collision or boundary exceedance that warrant close attention are marked in yellow, and the final planned trajectory for the ego vehicle is shown as green dots.
Fig.~\ref{fig:left_turn_vis}.~\ref{fig:waiting_for_pedestrian_vis}.~\ref{fig:right_turn_vis} shows the visualization of some of our planning results.

\section{Conclusion}
In this work, we introduce our winning solution for the NuPlan challenge. We adopt a novel spatial-temporal heatmap representation for planning, along with a corresponding post-solver to ensure a final plan that is both safe and comfortable. Experimental results validate the effectiveness of every component, highlighting our method's aptitude for balancing the ego progress and safety while generating safe and comfortable trajectories. 

{\small
\bibliographystyle{ieee_fullname}
\bibliography{egbib}
}

\end{document}